\documentclass[manuscript,screen]{acmart}

\usepackage[T1]{fontenc}
\usepackage{pifont}
\usepackage{array}
\usepackage{tikz}
\usepackage{amsmath}
\usetikzlibrary{arrows.meta, positioning, calc, fit, backgrounds, shapes.misc}
\usepackage{dsfont}
\usepackage{tabularx}
\usepackage{array}
\usepackage{threeparttable}
\usepackage{booktabs}

\newcolumntype{Y}{>{\raggedright\arraybackslash\footnotesize}X}

\definecolor{ink}{HTML}{283044}
\definecolor{modelfill}{HTML}{ECECEC}
\definecolor{deny}{HTML}{9E2B25}
\definecolor{denyfill}{HTML}{F4E2E1}
\definecolor{ok}{HTML}{1F6B4C}
\definecolor{okfill}{HTML}{DEEFE6}
\definecolor{cardline}{HTML}{C4CCDC}

\AtBeginDocument{%
  }

\setcopyright{none}
\authorsaddresses{}

\begin{document}

%
%

\definecolor{ink}{HTML}{283044}
\definecolor{modelfill}{HTML}{ECECEC}   
\definecolor{deny}{HTML}{9E2B25}        
\definecolor{denyfill}{HTML}{F4E2E1}
\definecolor{ok}{HTML}{1F6B4C}          
\definecolor{okfill}{HTML}{DEEFE6}
\definecolor{cardline}{HTML}{C4CCDC}

\newcommand{\Cmt}{\mathcal{C}}
\newcommand{\cUp}{\mathord{\uparrow}}
\newcommand{\cDn}{\mathord{\downarrow}}
\newcommand{\cChg}{\mathord{\updownarrow}}

\usetikzlibrary{positioning,fit,backgrounds,calc,arrows.meta,decorations.pathreplacing,shapes.geometric,shapes.misc}

\definecolor{catBalance}{RGB}{31,119,180}
\definecolor{catAge}{RGB}{230,126,0}
\definecolor{catUtil}{RGB}{44,160,44}
\definecolor{catDelq}{RGB}{214,39,40}
\definecolor{catInq}{RGB}{148,103,189}
\definecolor{inkgray}{RGB}{95,95,95}

\title{Navigating Disagreement Among Criteria for Evaluating Explanation Faithfulness}









\author{Grant Shanklin}
\orcid{0009-0004-5744-7732}
\affiliation{%
  \institution{Max Planck Institute for Security and Privacy}
  \city{Bochum}
  \country{Germany}
}
\affiliation{%
  \institution{Yale University}
  \city{New Haven}
  \state{Connecticut}
  \country{USA}
}
\email{grant.shanklin@yale.edu}

\author{Jiří Němeček}
\orcid{0009-0005-0585-1642}
\affiliation{%
  \institution{Max Planck Institute for Security and Privacy}
  \city{Bochum}
  \country{Germany}}
\email{}

\author{Asia Biega}
\orcid{0000-0001-8083-0976}
\affiliation{%
  \institution{Max Planck Institute for Security and Privacy}
  \city{Bochum}
  \country{Germany}}
\email{asia.biega@mpi-sp.org}


\begin{abstract}
Faithfulness requires that algorithmic explanations accurately reflect the model’s behavior. Yet existing operationalizations of faithfulness can disagree about whether the same explanation is faithful, leaving developers and auditors to choose among conflicting evaluations. In this paper, we investigate disagreement among evaluation criteria and offer a practical path forward for evaluating explanation faithfulness.

We first identify how operationalizations of faithfulness differ. We introduce a faithfulness functional that locates their disagreements in what an explanation claims about the model, which comparisons bear on that claim, and how agreement across those comparisons should be aggregated, and we reconstruct existing evaluation criteria within it. We argue these reconstructions support understanding faithfulness as a common though incompletely specified relation between the model and explanation, so the differences reflect choices left open by the general requirement. Justifying an operationalization then comes down to justifying these choices, which, on our view, draws on explanatory, institutional, and normative considerations. We support this argument with a case study of adverse action notices, reconstructing the evaluation implicit in an industry-proposed procedure and identifying which of its choices the setting justifies and which remain open.

Our work gives auditors and practitioners a way to navigate disagreeing evaluations of faithfulness. It also explains why the literature should expect multiple operationalizations, while showing how concrete settings can narrow the choice among them. More broadly, our analysis shows how a shared requirement of faithfulness can support contextually justified evaluations despite disagreement among existing metrics.
\end{abstract}




\maketitle

\section{Introduction}

Algorithmic explanations are commonly evaluated by breaking down desirable properties of explanations into distinct requirements and creating evaluations for each \cite{nautaAnecdotalEvidenceQuantitative2023}. These requirements are derived from the purpose the explanation is meant to serve and the context in which it is used \cite{luGoodExplanationAlgorithmic2020}. One such requirement is \textit{faithfulness}: the extent to which an explanation accurately represents the model’s behavior for a particular input \cite{jacoviFaithfullyInterpretableNLP2020, dasguptaFrameworkEvaluatingFaithfulness2022}.

Faithfulness is required for most use cases of explainability \cite{krutskyXAIDesiderataTrustworthy}, particularly when an explanation is used as evidence about why a model behaved as it did. For instance, faithfulness is required when a clinician uses an explanation to determine whether a model's diagnosis rests on medically relevant evidence; an unfaithful explanation could therefore induce reliance on the wrong grounds \cite{bussoneRoleExplanationsTrust2015}. Or a developer might use an explanation of an erroneous prediction to identify and correct the model's reliance on a spurious feature; here, unfaithfulness could obscure the source of the error and misdirect the attempted correction \cite{ribeiroWhyShouldTrust2016, adebayoSanityChecksSaliency2018}.

The requirement of faithful explanations has further legal grounding. In the United States (US), when creditors take action against a consumer, such as denying a loan application, they are legally required to provide an explanation of why that action was taken in the form of an adverse action notice (AAN). AANs include the list of ``principal reasons'' the creditor acted against the consumer \cite{ecoa, fcra}. As black-box machine learning models have been introduced to these decision-making processes \cite{treasuryAI2024, aldasoroIntelligentFinancialSystem2024}, algorithmic explanation methods have been proposed to generate such notices \cite{AdverseActionNotices, nairExplainingAdverseActions2024}. Both the regulatory text and the purposes of the underlying legislation require that the stated reasons accurately reflect the basis of the model’s decision, i.e., that the explanation be faithful \cite{finreglabPolicyAnalysis, selbstIntuitiveAppealExplainable2018}.

Establishing that explanations ought to be faithful, however, does not yet tell us how to generate faithful explanations or determine whether a given explanation is faithful. Doing so requires \textit{operationalizing} the general requirement, translating its natural-language definition into formal, mathematical terms, such that measurement and evaluation can be performed \cite{jacobsMeasurementFairness2021}. In our case, operationalizing faithfulness is what allows us to evaluate the extent to which an explanation is faithful to a model. 

The algorithmic explanation literature contains many operationalizations of faithfulness. This plurality is evident across methods for generating faithful explanations (e.g., \cite{ribeiroWhyShouldTrust2016, ribeiroAnchorsHighPrecisionModelAgnostic2018, carterWhatMadeYou2019}) and for evaluating the faithfulness of a given explanation (e.g., \cite{deyoungERASERBenchmarkEvaluate2020, dasguptaFrameworkEvaluatingFaithfulness2022, agarwalOpenXAITransparentEvaluation2022}). However, these operationalizations substantially differ in what they require \cite{krishnaDisagreementProblemExplainable2024}; a single explanation may be judged faithful under one and unfaithful under another \cite{azzolinReconsideringFaithfulnessRegular2025, barrDisagreementProblemFaithfulness2023}.

Multiple, non-equivalent operationalizations of faithfulness are problematic when we do not have a way to negotiate between them. Real-world implementations of faithful algorithmic explanations must choose among operationalizations, and so require a way to compare, evaluate, and select among different operationalizations of the same requirement. Existing work documents their disagreement \cite{krishnaDisagreementProblemExplainable2024, barrDisagreementProblemFaithfulness2023} and unifies the explanation methods they evaluate \cite{covertExplainingRemovingUnified2021}, but does not say what would justify one operationalization over another for a given use. This has led to the sobering conclusion that some of these operationalizations ``are not reliable enough to be used in real scenarios'' \cite{miro-nicolauComprehensiveStudyFidelity2025} and that, at least in settings with graph neural networks, faithfulness may not be ``a sensible design goal'' \cite{azzolinReconsideringFaithfulnessRegular2025}.

The task of our paper is to provide a framework for justifying choices among non-equivalent operationalizations of faithfulness. We make explicit the substantive, and often contextual, choices that enter an evaluation and examine what can justify them. This allows us to distinguish choices supported by an explanation's setting from those that are not. Our contribution helps explain why conflicting evaluations exist within the literature and gives auditors and practitioners a way to assess their suitability for a particular use. More broadly, this work bears on the theoretical foundations of algorithmic explanation, which have been criticized as under-theorized \cite{moshkovitzPositionExplainabilityResearch2026,watsonConceptualChallengesInterpretable2022,afrooghExplainableAIXAI2026}. We follow recent work \cite{jacobsMeasurementFairness2021,costonValidityPerspectiveEvaluating2023,wallachEvaluatingGenerativeAI2024} that develops these foundations through measurement modeling.

\subsection{Contributions}

This work is structured around the idea that understanding \textit{how} faithfulness evaluations in the literature differ will help explain \textit{why} they differ. Clarifying points of disagreement between operationalizations guides what matters when choosing among them. We thus make our contribution in three parts.

First, \textbf{we introduce a common representation of faithfulness operationalizations for two classes of local explanation methods (Section~\ref{sec:uff})}. We propose a three-component functional representation that writes each operationalization in the same formal terms. We show that a varied set of faithfulness operationalizations can be reconstructed within the functional framework, differing only in what we call the \textit{faithfulness specification}. Under this reconstruction, each operationalization tests whether the model's behavior accords with the claim attributed to the explanation. 

Second, \textbf{we use the reconstructions to argue that faithfulness names a common but incompletely specified relation between an explanation and model behavior (Section~\ref{sec:interpreting-uff})}. The reconstructed criteria differ in substantive choices that remain open even when faithfulness is understood as accurately representing model behavior. We distinguish these specification choices from disagreements about the underlying concept and errors in measuring a fixed target. This interpretation explains how multiple operationalizations can express a shared requirement while yielding different evaluations.

Finally, \textbf{we identify explanatory, institutional, and normative considerations that bear on the justification of a faithfulness specification (Section~\ref{sec:justifying-specifications})}. We apply this framework to adverse action notices, reconstructing the evaluation implicit in a US bank's proposed reason-selection procedure, which ranks fixed-length notices by their mean B-Shap contribution. We then examine which choices receive support from the disclosure setting and which remain unresolved. The case shows how the framework can make an evaluation's commitments explicit and subject them to contextual scrutiny.

\subsection{Related Work}

Existing work has sought to clarify the differences between algorithmic explanation methods by representing them in common formal terms. These representations identify choices about how features are removed and their influence summarized \cite{covertExplainingRemovingUnified2021}, which neighborhood is used to approximate the model \cite{hanWhichExplanationShould2022}, and which reference distribution is used to compute an attribution \cite{merrickExplanationGameExplaining2020}. Related work on evaluation shows that the choice of operationalization can affect whether an explanation is judged faithful. For instance, \citet{azzolinReconsideringFaithfulnessRegular2025} parameterize faithfulness metrics for graph neural networks and recommend choosing parameters according to the task. DeepFaith brings ten metrics into a common framework for training an explainer \cite{guoLearningHighlyFaithful2025}. Our faithfulness functional follows this approach of making the choices within existing methods explicit, though we use the resulting representation to consider what these choices require of an explanation instead of asking which one is, in itself, correct.

Disagreement among operationalizations has also been investigated through empirical evaluations of faithfulness metrics. \citet{tomsettSanityChecksSaliency2020} apply reliability tests from psychometrics and find that saliency metrics do not measure the same notion of fidelity. \citet{chanComparativeStudyFaithfulness2022} compare metrics by how often they prefer faithful explanations to random ones, while other work develops further approaches to evaluating the metrics themselves \cite{hedstromMetaEvaluationProblemExplainable2023, liEvaluatingReadabilityFaithfulness2024, barrDisagreementProblemFaithfulness2023}. Problems in the evaluation procedure, including the out-of-distribution inputs created by feature removal, can also affect these evaluations \cite{haseOutofDistributionProblemExplainability2021, juLogicTrapsEvaluating2022, bluecherDecouplingPixelFlipping2024}. However, determining how reliably an operationalization measures a given requirement does not yet establish that it specifies the requirement appropriately. Our concern is with the prior question of what justifies the specification an evaluation adopts.

Understanding faithfulness as incompletely specified has a close precedent in \citet{jacoviAligningFaithfulInterpretations2021}. They describe the requirement as ``vague and incomplete,'' and propose completing it by requiring explanations to be faithful to the causal chain people attribute to them. We share the view that evaluating faithfulness depends on what an explanation is taken to claim about the model. Our contribution is to identify how these claims enter existing operationalizations and to consider what justifies attributing a particular claim to an explanation. 

Further, the role of context in evaluating explanations has been established more broadly in the algorithmic explanation literature. Existing work identifies contestable epistemic values within explanation methods \cite{hancox-liEpistemicValuesFeature2021}, connects evaluation to contexts of use \cite{liaoConnectingAlgorithmicResearch2022}, and argues for formally specifying the goals an explanation is intended to achieve \cite{haufeExplainableAINeeds2026}. The need to justify such choices is particularly consequential in legal settings, where ambiguity in post-hoc explanations can allow providers to select explanations that are difficult to contest \cite{bordtPostHocExplanationsFail2022, selbstIntuitiveAppealExplainable2018}. We build on this work by considering how contextual requirements can justify the components of a faithfulness specification. In the case of adverse action notices, we examine an existing reason-selection procedure using the form of the notice, the rules governing its content, and the purposes of disclosure. This identifies grounds for some evaluation choices while showing where the procedure's mathematical properties leave their justification unresolved.

\section{Operationalizing faithfulness}
\label{sec:operationalizing-faithfulness}

The sense of faithfulness we are concerned with is the extent to which an explanation accurately reflects the model's behavior for a particular input \cite{jacoviFaithfullyInterpretableNLP2020, dasguptaFrameworkEvaluatingFaithfulness2022}.\footnote{We use `faithfulness' specifically for the relation between a local explanation of a particular model output and the model behavior it purports to represent. Other uses of the term in the algorithmic explanation literature (e.g., \cite{turpinLanguageModelsDon2023, guidottiLocalRuleBasedExplanations2018a}) fall outside the scope of this paper. We also note that terms such as completeness \cite{nautaAnecdotalEvidenceQuantitative2023} and fidelity \cite{islamUnifiedFrameworkEvaluating2024} are used synonymously with faithfulness in the literature, so we include their operationalizations when they explicitly refer to the same requirement as faithfulness.} When considering operationalizations, we focus only on evaluations of local feature-attribution explanations and selected rationales. These explanations assign importance scores to input features or identify a subset of input features to explain the model's behavior for some specific input. We consider these two explanation forms together because (i) their faithfulness operationalizations disagree and (ii) the two forms are related in practice, since feature attributions are commonly used to select rationales. However, a similar analysis may extend to other explanation forms, including counterfactual explanations \cite{altmeyerFaithfulModelExplanations2024, chowdhuryLookingMirrorFaithful2025}. 

The algorithmic explanation literature offers multiple operationalizations of faithfulness within this scope that rank the same pair of explanations differently. This raises a practical problem for developers and auditors of algorithmic explanations. If different operationalizations rank the same explanations differently, then the choice of operationalization partly determines which explanation will be judged more faithful. That choice, therefore, cannot be treated as a neutral measurement decision, and an account is required to navigate between them. We illustrate this problem through three existing operationalizations under a definition of rank non-equivalence.

Let $f\colon\mathcal X\to\mathcal Y$ denote the prediction model, with inputs $x\in\mathcal X\subseteq\mathbb R^d$. An explanation explains the model's output at $x$, such as a predicted class or a decision, and faithfulness is a relation between the explanation and $f$. Evaluations, however, observe $f$ only through a real-valued score $s\colon\mathcal X\to\mathbb R$ derived from it, oriented so that higher values indicate stronger support for the output being explained. For a classification model, $s$ is the score the model assigns to the class it predicts for $x$, such as that class's probability or logit, and the class is held fixed when $s$ is evaluated at other inputs.\footnote{We restrict attention to scores with an orientation of this kind. Extending the signed-change criteria considered here to other regression settings would require specifying which changes in the output count in favor of the explanation.} Write $\phi\in\mathbb R^d$ for an attribution vector and $R\subseteq[d]=\{1,\ldots,d\}$ for a selected rationale. The attribution vector assigns an importance value to each feature, while a rationale singles out a set of features as important without values. A rationale may be provided directly or selected from an attribution vector by a declared rule, most commonly its $k$ highest-attributed features $S_k(\phi)$ \cite{deyoungERASERBenchmarkEvaluate2020, agarwalOpenXAITransparentEvaluation2022}. When a rationale criterion is applied to an attribution vector, we use this top-$k$ rule. We use $e$ to stand in for an explanation of either form.

We focus on disagreements that reverse a strict ranking of explanations. Specifically, two operationalizations are non-equivalent when they prefer opposite explanations of the same model output.

\begin{definition}[Rank non-equivalence]
\label{def:rank-non-equivalence}
Let $F$ and $F'$ be two operationalizations of faithfulness, $s$ a model score, $x$ an input, and $e$, $e'$ two distinct explanations of the model's output at $x$ that both $F$ and $F'$ can score. Write $e\succ_F e'$ when $F$ scores $e$ as more faithful than $e'$. We call $F$ and $F'$ \emph{rank-non-equivalent} if there exist such $s$, $x$, $e$, and $e'$ with $e\succ_F e'$ and $e'\succ_{F'} e$.
\end{definition}

Consider the following three operationalizations of faithfulness. 

\begin{itemize}
    \item \textbf{Explanation infidelity.}
    \citet{yehInfidelitySensitivityExplanations2019} evaluate how well the attribution vector locally approximates changes in the model's output. Under a specified distribution of input perturbations, they compare the linear combination of attributions weighted by the perturbation with the resulting change in the score.
    \begin{equation}
        \operatorname{INFD}(\phi;s,x,\mu_I)
        =
        \mathbb E_{I\sim\mu_I}
        \left[
            \left(
                \phi^\top I-\left[s(x)-s(x-I)\right]
            \right)^2
        \right].
        \label{eq:infidelity}
    \end{equation}
    Lower infidelity indicates greater faithfulness.

    \item \textbf{Sufficiency.}
    \citet{deyoungERASERBenchmarkEvaluate2020} evaluate whether a selected rationale $R$ contains enough information to support the model's prediction. Here, $s$ is the probability of the predicted class, and $x^R$ contains only the tokens selected by $R$. Sufficiency is
    \begin{equation}
        \operatorname{SUFF}(R;s,x)
        =
        s(x)-s(x^R).
        \label{eq:sufficiency}
    \end{equation}
    Lower sufficiency scores indicate greater faithfulness.

    \item \textbf{Comprehensiveness.}
    \citet{deyoungERASERBenchmarkEvaluate2020} also evaluate whether the selected rationale contains information on which the model's prediction depends. With $s$ defined as above, let $x^{\setminus R}$ be the input obtained by deleting the tokens selected by $R$. Comprehensiveness is
    \begin{equation}
        \operatorname{COMP}(R;s,x)
        =
        s(x)-s(x^{\setminus R}).
        \label{eq:comprehensiveness}
    \end{equation}
    Higher comprehensiveness indicates greater faithfulness.
\end{itemize}

We show in Appendix~\ref{app:non-equivalence} that the three are pairwise rank-non-equivalent. The non-equivalence of these operationalizations shows that choosing among them is not a neutral measurement choice and an account is required to make that choice. We begin such an account by first identifying the substantive differences in what the operationalizations are evaluating and in what ways they differ. The next section introduces a common representation that makes these differences explicit, making the first step towards providing our account for navigating such substantive measurement decisions. 

\section{Faithfulness functional}
\label{sec:uff}

In this section, we characterize how operationalizations of faithfulness structurally differ. To do so, we introduce a functional for representing a broad class of faithfulness operationalizations, showing how each differs only in its comparisons, assessments, and aggregation rules. The functional makes the differences between operationalizations explicit and comparable, providing the basis for asking what can justify choosing between them in a particular explanatory setting. We limit the scope of the functional to evaluations that assess a fixed model through its responses to specified input comparisons. Evaluations that intervene on the model itself, such as ROAR's feature-removal and retraining procedure \cite{hookerBenchmarkInterpretabilityMethods2019}, require additional structure beyond the input comparisons formalized here.

We represent a faithfulness operationalization through three components. The first is a distribution $\nu_{x,e}$ over the comparisons to be made. Each comparison $\omega$ determines a point $x'_\omega$ against which the explained input is compared. The model's response to the comparison point is, then, the pair of scores $s(x)$ and $s(x'_\omega)$. The second component is an assessment rule $\ell_{x,e}$, which records what the operationalization takes a favorable model response to be. Usually the assessment is a real number, oriented so that larger values are more favorable to the explanation. The third component is an aggregation rule $\mathcal A$, which combines the assessments over all comparisons into a single number. Most criteria take the expectation, $\mathcal A=\mathbb E$, and so report the mean assessment, while others take the Pearson correlation between the observed and predicted responses. We refer to a particular choice of these three components as a \emph{faithfulness specification} and label it $\Sigma$.

\begin{definition}[Faithfulness functional]
\label{def:unified-faithfulness}
Fix the score $s$, an instance $x\in\mathcal X$, and an explanation $e$ of either form. For a faithfulness specification $\Sigma=\langle\nu_{x,e},\ell_{x,e},\mathcal A\rangle$, the faithfulness of $e$ is
\begin{equation}
\label{eq:unified-faithfulness}
\operatorname{Faith}_{\Sigma}(e;x,s)
=
\mathcal A_{\omega\sim\nu_{x,e}}
\bigl[\ell_{x,e}(\omega;s)\bigr],
\end{equation}
the aggregate of the assessments over comparisons drawn from $\nu_{x,e}$, with larger values indicating greater faithfulness. We write $x'$ for $x'_\omega$ where no ambiguity arises.
\end{definition}

Additionally, we set admissibility conditions to keep comparison generation, model-response assessment, and aggregation distinct. These conditions constrain how an evaluation is reconstructed in our functional.

\begin{definition}[Admissibility conditions]
\label{def:admissible}
A specification $\Sigma=\langle\nu_{x,e},\ell_{x,e},\mathcal A\rangle$ is \emph{admissible} when
\begin{itemize}
    \item[(i)]$\nu_{x,e}$ makes no additional use of $s$ to select or reweight comparison inputs, given $x$, $e$, and declared reference information fixed independently of $e$;
    \item[(ii)]$\ell_{x,e}(\omega;s)$ uses $s$ only through $s(x)$ and $s(x'_\omega)$, and depends nontrivially on at least one of these values;
    \item[(iii)] $\mathcal A$ depends only on the distribution of assessments under $\nu_{x,e}$, and for scalar assessments it is monotone: raising the assessment of any comparison never lowers the aggregate;
    \item[(iv)] at least one of $\nu_{x,e}$ and $\ell_{x,e}$ depends on $e$.
\end{itemize}
\end{definition}

An admissible specification reconstructs an existing criterion when it preserves what Section~\ref{sec:operationalizing-faithfulness} showed criteria to disagree about, namely how they rank explanations. We therefore require a reconstruction to recover the source criterion $F$, negated where lower values are preferred, up to a positive rescaling and shift:
\[
\operatorname{Faith}_{\Sigma}(e;x,s)=a_{x,s}F(e;x,s)+b_{x,s},\qquad a_{x,s}>0,
\]
where the constants may depend on $x$ and $s$ but not on $e$. A reconstruction then ranks every pair of explanations exactly as its source does, so it can never be rank-non-equivalent to its source in the sense of Definition~\ref{def:rank-non-equivalence}. To divide a source's computation between the components, we take the assessment to be the quantity it computes for each comparison and the aggregation rule to be the operation it applies across comparisons. Where a source still admits more than one admissible reading, we attempt to match its own description with the interpretive meaning of each component of the faithfulness specification. However, substantive interpretation concerns the specification as a whole. Equivalent reconstructions may distribute a computation differently between assessment and aggregation, so the components provide a way to expose an evaluation's choices without requiring a unique decomposition.

To show how the functional expresses existing operationalizations, we reconstruct the infidelity of Equation~\eqref{eq:infidelity} \cite{yehInfidelitySensitivityExplanations2019}, where $e=\phi$ is a feature attribution vector. First, the comparison distribution. In the original paper, a perturbation $I$ is drawn from $\mu_I$, and the point $x-I$ is compared with $x$. A comparison is therefore the draw $\omega=I\sim\mu_I$, with comparison point $x'=x-I$. This assignment does not consult the model, satisfying (i). Second, the assessment rule. The attribution predicts the score change $\phi^\top I$, equivalently $\phi^\top(x-x')$, a quantity read off the two points being compared. The source penalizes the squared discrepancy between the prediction and the observed change, so
\[
\ell^{\operatorname{INFD}}_{x,\phi}(\omega;s)
=
-\Bigl(
s(x)-s(x')-\phi^\top(x-x')
\Bigr)^2 .
\]
This uses $s$ only through $s(x)$ and $s(x')$ and varies with both, satisfying (ii), and it depends on $\phi$, satisfying (iv). Third, the aggregation rule is the expectation, $\mathcal A^{\operatorname{INFD}}=\mathbb E$, which depends only on the distribution of assessments and is monotone, satisfying (iii). The specification is therefore admissible, and
\[
\operatorname{Faith}_{\Sigma^{\operatorname{INFD}}}(\phi;x,s)
=
-\mathbb E_{I\sim\mu_I}
\left[
\left(
\phi^\top I-\left[s(x)-s(x-I)\right]
\right)^2
\right]
=
-\operatorname{INFD}(\phi;s,x,\mu_I).
\]

Since INFD prefers lower values (see Section~\ref{sec:operationalizing-faithfulness}), we must orient infidelity as $-\operatorname{INFD}$ to match the ranking order. This is then a reconstruction with $a_{x,s}=1$ and $b_{x,s}=0$.

The value of the faithfulness functional lies in making the choices of varied operationalizations explicit and comparable, so that each can be justified. This can only be shown by representing varied operationalizations, specifying precisely how they are similar and different. Thus, we select prominent operationalizations identified in the algorithmic explanation literature, prioritizing variation in the model comparisons they examine and the criteria they apply.

Table~\ref{tab:unified-faithfulness} reconstructs eight criteria in this way. Appendix~\ref{app:translation-long-notes} gives the worked translations and the constants in each case. This shows a general pattern: the operationalizations alter either the features an explanation identifies or the remaining features, at one set size or across several, and score the resulting change in the model’s output.

\begin{table*}[t]
  \centering
  \begin{threeparttable}

  \caption{
      This table reconstructs eight operationalizations of faithfulness using the proposed faithfulness functional. Read left to right by row, the first column introduces the operationalization, with a short description of what it evaluates. The second column is the set of comparisons $\omega$, which specifies how comparison inputs $x'$ are constructed or sampled. The third column is the assessment function $\ell$, which assesses the model's response at each comparison point. The fourth column is the aggregation method $\mathcal{A}$, which combines the assessments into a single score. The faithfulness specification in each row produces a reconstruction, when put into the faithfulness functional, that preserves its source's ranking of explanations. We give derivations for each in Appendix~\ref{app:translation-long-notes}. Comparing the rows shows how operationalizations differ across specification parts, yielding three broad categories that we use to organize the table: correspondence, where attribution values should predict or track observed score changes; dependence, where disturbing the identified features should move the score; and retained support and invariance, where the selected information should support a high score or perturbing the remaining features should leave it unchanged.  
    }
  \label{tab:unified-faithfulness}

  \footnotesize
  \setlength{\tabcolsep}{2.6pt}
  \renewcommand{\arraystretch}{1.10}

  \renewcommand{\tabularxcolumn}[1]{m{#1}}

  \newcommand{\opentry}[3]{%
    \textbf{#1}~\citep{#2}\par
    \vspace{0.5pt}%
    {\scriptsize\itshape
      \linespread{0.95}\selectfont
      #3\par
    }%
  }

  \begin{tabularx}{\textwidth}{
    @{}
    >{\raggedright\arraybackslash}m{4.2cm}
    >{\centering\arraybackslash}m{3.6cm}
    >{\centering\arraybackslash}X
    >{\centering\arraybackslash}m{1.75cm}
    @{}
  }

    \toprule

    Operationalization
      &
    Comparison $\omega\sim\nu_{x,e}$
      &
    Assessment $\ell_{x,e}(\omega;s)$
      &
    Aggregation $\mathcal A$
    \\

    \midrule

    \multicolumn{4}{@{}p{\textwidth}@{}}{%
      \itshape
      Correspondence: attribution values should predict or track observed score changes.
    }
    \\[2pt]

    \opentry
      {Explanation infidelity}
      {yehInfidelitySensitivityExplanations2019}
      {The attribution vector should predict the score change associated with each perturbation.}
      &
    $\displaystyle
      \begin{gathered}
        I\sim\mu_I,\\[-1pt]
        x'=x-I
      \end{gathered}
    $
      &
    $\displaystyle
      -\Bigl(
        s(x)-s(x')-\phi^\top(x-x')
      \Bigr)^2
    $
      &
    $\displaystyle \mathbb E$
    \\
    \addlinespace[3pt]

    \opentry
      {Faithfulness correlation}
      {3491440.3491857}
      {Attribution sums should track the score changes caused by replacing feature subsets.}
      &
    $\displaystyle
      \begin{gathered}
        S\sim\operatorname{Unif}
        \{T\subseteq[d]:|T|=n\},\\[-1pt]
        x'=x^{\setminus S}
      \end{gathered}
    $
      &
    $\displaystyle
      \Bigl(
        s(x)-s(x'),
        \;
        \textstyle\sum_{j\in S}\phi_j
      \Bigr)
    $
      &
    $\displaystyle \operatorname{Corr}$
    \\

    \addlinespace[2pt]
    \midrule
    \addlinespace[1pt]

    \multicolumn{4}{@{}p{\textwidth}@{}}{%
      \itshape
      Dependence: disturbing the identified features should move the score.
    }
    \\[2pt]

    \opentry
      {Comprehensiveness}
      {deyoungERASERBenchmarkEvaluate2020}
      {Removing the selected rationale should suppress the model score.}
      &
    $\displaystyle x'=x^{\setminus R}$
      &
    $\displaystyle
      s(x)-s(x')
    $
      &
    $\displaystyle \mathbb E$
    \\
    \addlinespace[3pt]

    \opentry
      {AOPC}
      {samekEvaluatingVisualizationWhat2017}
      {Removing earlier-ranked features should rapidly suppress the model score.}
      &
    $\displaystyle
      \begin{gathered}
        S\sim\pi_\phi,\\[-1pt]
        x'\sim q \bigl(\,\cdot\mid x,S\bigr)
      \end{gathered}
    $
      &
    $\displaystyle
      s(x)-s(x')
    $
      &
    $\displaystyle \mathbb E$
    \\
    \addlinespace[3pt]

    \opentry
      {PGI}
      {agarwalOpenXAITransparentEvaluation2022}
      {Perturbing the highest-ranked features should change the prediction substantially.}
      &
    $\displaystyle
      \begin{gathered}
        S\sim\pi_\phi,\\[-1pt]
        x'\sim q
        \bigl(\,\cdot\mid x,S\bigr)
      \end{gathered}
    $
      &
    $\displaystyle
      \bigl| s(x)-s(x') \bigr|
    $
      &
    $\displaystyle \mathbb E$
    \\

    \addlinespace[2pt]
    \midrule
    \addlinespace[1pt]

    \multicolumn{4}{@{}p{\textwidth}@{}}{%
      \itshape
      Retained support and invariance: selected information should support a high score, or perturbing the remaining features should leave it unchanged.
    }
    \\[2pt]

    \opentry
      {Sufficiency}
      {deyoungERASERBenchmarkEvaluate2020}
      {Retaining only the selected rationale should retain a high original-class score.}
      &
    $\displaystyle x'=x^{R}$
      &
    $\displaystyle
      s(x')-s(x)
    $
      &
    $\displaystyle \mathbb E$
    \\
    \addlinespace[3pt]

    \opentry
      {Insertion AUC}
      {petsiukRISERandomizedInput2018}
      {Restoring earlier-ranked features should rapidly recover a high original-class score.}
      &
    $\displaystyle
      \begin{gathered}
        S\sim\pi_\phi,\\[-1pt]
        x'=x^{S}
      \end{gathered}
    $
      &
    $\displaystyle
      s(x')
    $
      &
    $\displaystyle \mathbb E$
    \\
    \addlinespace[3pt]

    \opentry
      {PGU}
      {agarwalOpenXAITransparentEvaluation2022}
      {Perturbing features outside the highest-ranked set should leave the prediction unchanged.}
      &
    $\displaystyle
      \begin{gathered}
        S\sim\pi_\phi,\\[-1pt]
        x'\sim q
        \bigl(\,\cdot\mid x,[d]\setminus S \bigr)
      \end{gathered}
    $
      &
    $\displaystyle
      -\bigl| s(x)-s(x') \bigr|
    $
      &
    $\displaystyle \mathbb E$
    \\

    \bottomrule

  \end{tabularx}

  \begin{tablenotes}[flushleft]
  \footnotesize
  \item \textit{Comparison Notation.}
  $x^{\setminus S}$ denotes an input with the features in $S$
  removed or replaced; $x^S$ denotes an input with those features
  retained or restored. The precise operation depends on the
  criterion. The same notation applies to a rationale $R$.
  $\operatorname{Unif}\{T\subseteq[d]:|T|=n\}$ selects uniformly
  among all subsets of a fixed size $n$.
  $S\sim\pi_\phi$ selects among the top-ranked sets $S_k(\phi)$
  evaluated by the criterion, with probabilities given by its
  normalized averaging or integration weights. The evaluated
  sizes and weights are fixed across explanations; the selected
  features depend on $\phi$.
  $q(\cdot\mid x,S)$ perturbs features in $S$ while holding the
  others fixed; $[d]\setminus S$ denotes the complementary set.

  \item \textit{Assessment Notation.}
  Each entry is a scalar assessment, except in faithfulness
  correlation, where the ordered pair contains the observed
  score drop and the sum of attributions over the selected features.

  \item \textit{Aggregation Notation.}
  $\mathbb E$ takes the mean assessment over the comparisons.
  $\operatorname{Corr}$ computes the Pearson correlation between
  the paired quantities, requiring both to have positive variance.
\end{tablenotes}

  \end{threeparttable}

\end{table*}

\section{Interpreting the faithfulness functional}
\label{sec:interpreting-uff}

\begin{figure}[tb]
\centering
\resizebox{\linewidth}{!}{%
\begin{tikzpicture}[
  x=1cm,y=1cm,
  font=\small,
  box/.style={draw=black!55,rounded corners=2pt,fill=white,
    minimum width=0.72cm,minimum height=0.48cm,inner sep=2pt},
  flow/.style={-{Stealth[length=1.6mm]},draw=black!65,semithick},
  heading/.style={align=center,font=\small\bfseries,text width=3.0cm},
  status/.style={align=center,font=\scriptsize,text width=3.0cm}
]
\node[anchor=east,align=right] at (0,0) {Theoretical\\understanding};
\node[anchor=east] at (0,-1.0) {Specification};
\node[anchor=east] at (0,-2.0) {Operationalization};

\node[heading] at (1.85,1.05) {(a) Theoretical\\contestation};
\node[box] (aC1) at (1.10,0) {$C_1$};
\node[box] (aC2) at (2.60,0) {$C_2$};
\node[box] (aS1) at (1.10,-1.0) {$\Sigma_1$};
\node[box] (aS2) at (2.60,-1.0) {$\Sigma_2$};
\node[box] (aO1) at (1.10,-2.0) {$O_1$};
\node[box] (aO2) at (2.60,-2.0) {$O_2$};
\draw[flow] (aC1)--(aS1); \draw[flow] (aC2)--(aS2);
\draw[flow] (aS1)--(aO1); \draw[flow] (aS2)--(aO2);
\node[status,text=black!65] at (1.85,-2.65) {Alternative interpretation};

\node[heading] at (5.35,1.05) {(b) Incomplete\\specification};
\node[box,draw=teal!65!black] (bC) at (5.35,0) {$C$};
\node[box,draw=teal!65!black] (bS1) at (4.60,-1.0) {$\Sigma_1$};
\node[box,draw=teal!65!black] (bS2) at (6.10,-1.0) {$\Sigma_2$};
\node[box,draw=teal!65!black] (bO1) at (4.60,-2.0) {$O_1$};
\node[box,draw=teal!65!black] (bO2) at (6.10,-2.0) {$O_2$};
\draw[flow] (bC)--(bS1); \draw[flow] (bC)--(bS2);
\draw[flow] (bS1)--(bO1); \draw[flow] (bS2)--(bO2);
\node[status,font=\scriptsize\bfseries,text=teal!65!black]
  at (5.35,-2.65) {Interpretation defended here};

\node[heading] at (8.85,1.05) {(c) Different indicators\\of a fixed specification};
\node[box] (cC) at (8.85,0) {$C$};
\node[box] (cS) at (8.85,-1.0) {$\Sigma$};
\node[box] (cO1) at (8.10,-2.0) {$O_1$};
\node[box] (cO2) at (9.60,-2.0) {$O_2$};
\draw[flow] (cC)--(cS);
\draw[flow] (cS)--(cO1); \draw[flow] (cS)--(cO2);
\node[status,text=black!65] at (8.85,-2.65) {Alternative interpretation};

\begin{scope}[on background layer]
  \filldraw[fill=teal!5,draw=teal!55!black,rounded corners=4pt]
    (3.70,-2.98) rectangle (7.00,1.52);
\end{scope}
\end{tikzpicture}%
}
\caption{Three interpretations of multiple faithfulness operationalizations. Plurality may originate in (a) competing theoretical understandings, (b) different specifications of a shared but incomplete requirement, or (c) different indicators of one fixed specification. $C$ denotes a theoretical understanding, $\Sigma$ a faithfulness specification, and $O$ an operationalization. The three are not exclusive; we argue that (b) best explains the substantive differences reconstructed by the faithfulness functional. Two branches are shown for illustration. The distinction among levels is adapted from the measurement validity literature \cite{adcockMeasurementValidityShared2001, jacobsMeasurementFairness2021, jacobsMeasurementGovernanceResponsible2021, wallachEvaluatingGenerativeAI2024}.}
\label{fig:measurement-options}
\end{figure}
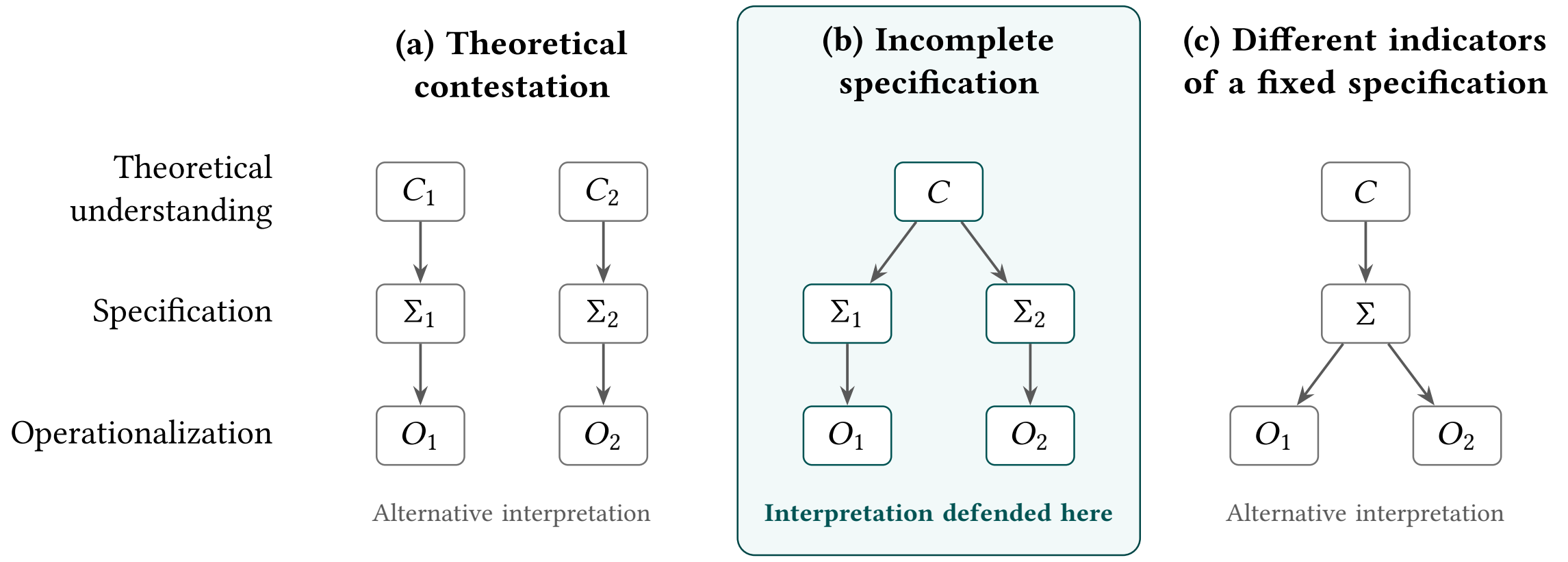

In this section, we use the reconstructions to argue that faithfulness names a common, incompletely specified relation between model and explanation. Drawing on the measurement modeling literature \cite{adcockMeasurementValidityShared2001,jacobsMeasurementFairness2021,wallachEvaluatingGenerativeAI2024}, we distinguish three sources of disagreement among operationalizations, shown in Figure~\ref{fig:measurement-options}. We argue that the reconstructed criteria expose substantive choices at the level of specification and then give an independent argument for why faithfulness leaves such choices open.

The first possible source of disagreement is the theoretical understanding of the concept (shown as (a) in Figure~\ref{fig:measurement-options}). This corresponds to disagreement about the background concept being measured \cite{adcockMeasurementValidityShared2001, wallachEvaluatingGenerativeAI2024}. Some concepts, such as fairness \cite{jacobsMeasurementFairness2021} and privacy \cite{mulliganPrivacyEssentiallyContested2016}, are contested at this level because the broad concept has rival, evaluative conceptions of what the concept really means \cite{collierEssentiallyContestedConcepts2006, gallieEssentiallyContestedConcepts1955, machery100YearsPsychology2007}. However, these rival conceptions usually do share a general description, so the fact that every operationalization we consider concerns accurately representing model behavior does not rule out contestation. For faithfulness, the rival conceptions would disagree about which kinds of facts about the model an explanation must match to be faithful. For instance, one account might say an explanation is faithful when the model's internal computation runs through the explanation, while another could hold that the explanation is faithful when removing the explanation lowers the model's score. Importantly, these contest the standard of accuracy itself, not the procedure for verifying an explanation's accuracy. A conception of faithfulness fixes which kinds of evidence about the model bear on faithfulness at all.
 
The second possible source of disagreement is at the level of the specification of a general requirement for use in a particular setting (shown as (b) in Figure~\ref{fig:measurement-options}). Specification is the step that brings in the goals and context of the inquiry to formalize precisely what the background concept means \cite{adcockMeasurementValidityShared2001}. Disagreement at the level of specification happens when a background concept can support several such formulations, each particular to the setting, not claiming to exhaust the concept's meaning \cite{adcockMeasurementValidityShared2001, wallachEvaluatingGenerativeAI2024}. The task at this level is to justify which formulation fits a particular use. For faithfulness, if the background concept requires that an explanation make an accurate claim about the model's behavior, then specification would involve making precise what that claim is and what the scope of such a claim is in that particular setting. Differences in specifications, if disagreement occurs at this level, must reflect disputes that persist after a conception of the background concept is fixed.
 
The third possible source of disagreement is different indicators for a fixed specification (shown as (c) in Figure~\ref{fig:measurement-options}). Here, operationalizations differ in the measurement procedure, while agreeing on the prior question of what to measure. Each procedure yields an indicator, and disagreements come down to method-specific errors in measuring the same quantity \cite{campbellConvergentDiscriminantValidation1959}. Such disagreement is, then, resolved by validation, for example by checking whether the indicators converge \cite{campbellConvergentDiscriminantValidation1959, adcockMeasurementValidityShared2001}. One example of indicator variation for faithfulness is disagreement about how to approximate an agreed-upon comparison distribution. 
 
We argue that the reconstructions within the faithfulness functional most closely match the type of disagreement predicted at the level of specification. In every reconstruction, an explanation is accurate or inaccurate according to the model's scores at $x$ and at the comparison points (condition (ii) of Definition~\ref{def:admissible}). Against this shared standard of accuracy, the functional locates the differences among operationalizations in what the explanation claims about the model ($\ell_{x,e}$), the comparison points over which that claim is evaluated ($\nu_{x,e}$), and how those comparisons are aggregated into a single judgment ($\mathcal A$). Each of these concerns the content or scope of the explanation's claim. These are the disputes that specification predicts will persist once the standard of accuracy is fixed, that is, once it is settled that faithfulness concerns whether an explanation's claim about the model is correct. They are also choices a setting may settle, since what an explanation claims depends on what it is used to assert. Section~\ref{sec:justifying-specifications} shows this for adverse action notices. 

This reading also explains why the disagreement is neither contestation nor indicator variation. Against (a), each reconstruction shares an understanding of faithfulness as being concerned with whether an explanation's claim about a model is correct through testing that claim against comparison points. In this sense, the faithfulness specification shows that the operationalizations share a standard of accuracy for measuring faithfulness, even though how that standard is tested varies. Against (c), each reconstruction recovers its source criterion exactly, so the differences between reconstructions are differences in the quantity each criterion defines, not in how well each computes it. For instance, the distribution $\nu_{x,e}$ differs across specifications. Indicator variation would instead predict a shared distribution, approximated by different approaches that introduce indicator errors \cite{haseOutofDistributionProblemExplainability2021, juLogicTrapsEvaluating2022}. However, we leave open the possibility that the specification does not fully determine the operationalization, with some choices, such as the reference against which features are removed, falling on either side of this line depending on the setting and method. 

To further bolster our case, we briefly argue that disagreement at the level of specification also follows from the requirement itself, independently of the reconstructions. Faithfulness, being a form of accuracy, can only be assessed against a definite claim. However, the explanation forms we consider do not fix their own claim. A rationale can be read as claiming that the prediction depends on its features or that its features suffice to support it. An attribution vector can be read as predicting how the score changes when each feature is perturbed, as ranking features by importance, or, through its highest entries, as selecting a set of features. The general requirement is that the explanation be accurate, yet it leaves open the specific claim the explanation makes; the explanation's purpose and use fill this gap, considerations brought in only at the time of specification. The general requirement therefore cannot be complete on its own, and the reconstructions show that the disagreement in the literature lies where this incompleteness predicts.

Both interpreting the reconstructions in Table~\ref{tab:unified-faithfulness} in light of the taxonomy from measurement theory and the independent argument support our primary conceptual claim: \textit{faithfulness is a requirement that names a common, incompletely specified relation between a model and an explanation.} The shared requirement fixes the standard of accuracy, namely the model's behavior, but leaves open what the explanation claims, which alternatives the claim concerns, and how strictly it must hold. This explains why the literature contains multiple operationalizations, while still allowing determinate judgments of faithfulness once a specification is fixed. It also changes what disagreement among operationalizations calls for. Such disagreement is not resolved by finding the correct conception of faithfulness or the most accurate indicator, but by asking which specification fits the explanation's use. This conclusion holds at least within our scope of faithfulness measures for prediction models being explained with feature attributions or rationales. The question of how to navigate disagreement among evaluation criteria then turns on how to justify choosing one specification over another in a concrete setting. 

\section{Justifying a faithfulness operationalization}
\label{sec:justifying-specifications}

In this section, we show that the faithfulness functional can be used to show which parts of faithfulness require specification and consider what would justify such a choice. We argue that three kinds of considerations bear: \emph{explanatory} considerations concern the form of the explanation and what it explains; \emph{institutional} considerations concern the rules governing what it must disclose; and \emph{normative} considerations concern the purposes the disclosure serves. These considerations can provide grounds for selecting comparisons, assessments, and aggregation in a faithfulness specification. To make this case, we work through adverse action notices generated by the procedure of \citet{nairExplainingAdverseActions2022}. We select this case study because it was proposed by industry rather than as a purely academic method. Researchers at the US bank Wells Fargo proposed the procedure, and the company later filed a patent application for it. We reconstruct an evaluation that recovers their selection of reasons and assess whether the setting justifies the choices this reconstruction reveals. 

\subsection{Adverse action notices}
\label{sec:intro-aans}

In the US, a creditor that takes adverse action against a consumer must provide an AAN stating the ``principal reasons'' for that action \cite{ecoaRegs, ecoa}. Originating in the 1976 amendments to the Equal Credit Opportunity Act (ECOA) and implemented through Regulation B, this requirement applies to credit denials and certain unfavorable changes to or refusals to increase existing credit \cite{ecoa, ecoaRegs}. Creditors have long used scorecards and other simple algorithms to evaluate credit applicants \cite{reserve2007report}. Recently, this has extended to the use of more complex machine learning models \cite{barocasHiddenAssumptionsCounterfactual2020, aldasoroIntelligentFinancialSystem2024}.\footnote{The regulatory authority, the Consumer Financial Protection Bureau (CFPB), clarified in 2022 that the obligation to provide accurate reasons applies undiminished to decisions made by complex algorithms \cite{ConsumerFinancialProtection2024, ConsumerFinancialProtection2023-03}. These circulars were withdrawn in 2025 as part of a broader withdrawal of CFPB guidance \cite{WithdrawnGuidance2026}. A 2026 final rule removed the effects test from Regulation B but left the notification requirements and the prohibition of disparate treatment in place \cite{cfpbRegB2026}. The statutory requirements and the official interpretation of Regulation B on which we rely were not affected.}

The official interpretation clarifies that a creditor need not describe how or why a factor adversely affected an applicant (comment 9(b)(2)-3), and no single method for selecting reasons is required \cite{cfpbOfficialInterp10029Notifications}. Two example methods identify factors on which the applicant scored furthest below the corresponding averages among applicants at or slightly above the passing score, or among all applicants (comment 9(b)(2)-5) \cite{cfpbOfficialInterp10029Notifications}. These examples provide concrete benchmarks for identifying principal reasons while allowing different approaches to their selection.

The requirement that AANs be faithful to the decision model derives from the regulatory text. The official interpretation states that the reasons disclosed must ``relate to and accurately describe the factors actually considered or scored by a creditor'' (comment 9(b)(2)-2) \cite{cfpbOfficialInterp10029Notifications}. For credit scoring systems, the disclosed reasons must ``relate only to those factors actually scored,'' and ``no factor that was a principal reason for adverse action may be excluded from disclosure'' (comment 9(b)(2)-4) \cite{cfpbOfficialInterp10029Notifications}. These are interpreted as imposing a model-grounded accuracy requirement \cite{finreglabPolicyAnalysis} that falls within the sense of faithfulness examined in this paper.

Further support for this requirement can be found in the antidiscriminatory purpose of adverse action notices. Legislative history says that AANs are designed to discourage discriminatory lending by forcing creditors to state the basis of their decisions, on the theory that creditors who must explain themselves are less likely to discriminate \cite{selbstIntuitiveAppealExplainable2018}. ``Only if creditors know they must explain their decisions, will they effectively be discouraged from discriminatory practices'' \cite{ecoaSenateReport}. This mechanism principally targets direct disparate treatment \cite{gillisInputFallacy2021, barocasBigDatasDisparate2016}. An explanation intended to discourage such discrimination must faithfully represent the grounds of the decision; otherwise, a creditor could disclose an innocuous reason even when a protected characteristic was the actual basis for the denial.

AANs, however, have additional purposes that turn out to bear on what a faithful explanation must disclose. Specifically, AANs' educational purpose requires that the notice allow consumers to understand where their application fell short, so that they can improve it or correct errors \cite{ecoaSenateReport, finreglabPolicyAnalysis, selbstIntuitiveAppealExplainable2018}. This impacts what an explanation is licensed to be interpreted as claiming. Both purposes therefore place demands on what it means for a notice to accurately identify the principal reasons for a decision, as we examine in the following sections.

\subsection{Reconstructing Nair et al.’s reason selection}
\label{sec:nair-reconstruction}
 
We take the procedure of \citet{nairExplainingAdverseActions2022}, proposed by Wells Fargo researchers to generate AANs and later the subject of a patent application by the company \cite{nairExplainingAdverseActions2024}, as our primary example of an algorithmic explanation procedure proposed for generating AANs.\footnote{Other proposed AAN tools \cite{AdverseActionNotices, h2oAIUnderstandingModelInterpretation2026, zestAIModelExplainabilityRe, shanbhagUnderstandingCounterfactualAttribution2020, krivorotovExplainingDenialsAdverse2022} are noted but not taken up, as they provide less detail about their procedures. \citet{finRegLabExplainabilityAndFairness2023} test such tools using different criteria for assessing the reasons disclosed.} The method groups related features into factors that correspond to reasons communicable to the applicant; for example, several measures of credit history may form one factor. Baseline Shapley (B-Shap) \cite{sundararajanManyShapleyValues2020} divides the difference between the applicant's score and that of an approved reference profile among these factors. It computes each factor's contribution by averaging its marginal effect over all orders in which factors can switch from the reference profile to the applicant's values. The factors with the largest contributions are then selected as the principal reasons communicated to the applicant \cite{nairExplainingAdverseActions2022}.
 
Selecting the $k$ factors with the largest B-Shap contributions is equivalent to maximizing the mean contribution of the factors named in a notice. We reconstruct this ordering within the faithfulness functional to examine the evaluation choices it expresses. The reconstruction preserves the ranking of every pair of fixed-length notices, allowing us to ask what would justify that ranking as an evaluation of whether a notice identifies the principal reasons for denial.
 
Here $f(z)\in(0,1)$ predicts default probability, and the creditor denies when $f(z)>\tau$.\footnote{We consider denials made by the model and its threshold alone, setting aside human intervention in the decision pipeline.} Following \citet{nairExplainingAdverseActions2022}, we work with the log-odds $s(z)=\operatorname{logit}f(z)$, which is higher for less favorable applications and supports the denial being explained. We fix the number of reasons at $k$ and compare notices naming that many factors.
 
\begin{definition}[A faithfulness specification reconstructing B-Shap notice rankings]\label{def:nair}
Fix a denied applicant $x$, a partition $\mathcal G$ of the features $[d]$ into $m$ factors, the approved reference $b$, and a notice length $1\leq k<m$. A notice $N\subseteq\mathcal G$ names $k$ factors, with rationale $R_N=\bigcup_{g\in N}g$. For a group of factors $D\subseteq\mathcal G$, let $x^{\setminus D}$ replace the features in those factors by their values in $b$, leaving the rest of the application fixed. Let $\kappa$ be the Shapley-kernel distribution over groups of factors,
\[
\kappa(D)\propto\frac{1}{\binom m{|D|}\,|D|\,(m-|D|)},
\qquad \varnothing\ne D\subsetneq\mathcal G.
\]
The specification $\Sigma^{\operatorname{Nair}}=\langle\nu^{\operatorname{Nair}}_{x,R_N},\ell^{\operatorname{Nair}}_{x,R_N},\mathcal A^{\operatorname{Nair}}\rangle$ draws comparisons $\omega=D\sim\kappa$ with $x'=x^{\setminus D}$, and sets
\[
\ell^{\operatorname{Nair}}_{x,R_N}(\omega;s)=\Bigl(s(x)-s(x'),\;\frac{|N\cap D|}{k}\Bigr),
\qquad
\mathcal A^{\operatorname{Nair}}=\operatorname{Corr}.
\]
A notice $N$ therefore has faithfulness score
\begin{equation}
\label{eq:aan-correlation}
\operatorname{Faith}_{\Sigma^{\operatorname{Nair}}}(R_N;x,s)
=\operatorname{Corr}_{D\sim\kappa}\left(
s(x)-s(x^{\setminus D}),\;
\frac{|N\cap D|}{k}
\right).
\end{equation}
\end{definition}
 
This specification represents the notice ranking through a variant of faithfulness correlation. It rewards association between larger score reductions and replacement of a larger fraction of the notice's named factors. Appendix~\ref{app:aan-correlation} derives the Shapley-kernel weighting by rewriting B-Shap contributions in terms of these score reductions and shows that, for fixed notice length, the correlation score is a positive affine transformation of the mean B-Shap contribution of the named factors.\footnote{Pearson correlation requires positive variances. The appendix treats the zero-variance case by assigning all notices the same score, preserving their tie.} The specification therefore preserves all rankings and ties among the candidate notices, and its maximizers are exactly those selected by the procedure. We call this the allocation-based ranking, since it ranks notices by the share of the applicant--reference score difference that B-Shap allocates to their factors. We use this representation to examine whether it is justified for AANs.
 
\subsection{Justifying the specification}
\label{sec:aan-justification}
 
We now consider how explanatory, institutional, and normative considerations justify the way $\Sigma^{\operatorname{Nair}}$ fills out a faithfulness specification for AANs. We focus on its comparison distribution $\nu^{\operatorname{Nair}}_{x,R_N}$ because it sets both the reference $b$ toward which the applicant's factors are replaced and the Shapley kernel $\kappa$, which determines which factors are replaced together. These provide two rich questions to test against the three kinds of considerations. The assessment and aggregation also require justification, but we leave them for future work. We argue that the setting justifies filling out the reference with a benchmark tied to the standard for approval, which Nair et al.'s procedure permits but their illustration does not adopt. For interactions, we identify competing considerations bearing on the allocation-based ranking and explain what further justification its use for disclosure requires.
 
Considering the reference, each comparison in $\nu^{\operatorname{Nair}}_{x,R_N}$ replaces some of the applicant's factors with their values in $b$. Thus, the choice of $b$ determines the contrast against which the notice's claim is evaluated. Explanatory considerations favor tying this contrast to the standard for approval, since an AAN explains why an application was denied rather than approved. The first regulatory example supports this institutionally, as it compares the applicant's factor scores with their averages among applicants at or slightly above the passing score (comment 9(b)(2)-5) \cite{cfpbOfficialInterp10029Notifications}. Normative considerations also support choosing an approved applicant as the reference because the educational purpose of the AAN requires the notice to direct applicants to where their application fell short of qualifying. More specifically, the educational purpose favors a reference closer to the decision boundary, rather than an applicant well within the accepted zone. A reference near the decision boundary in the direction of the denied applicant ties the comparison more closely to the standard the applicant failed to meet. 
 
Nair et al.'s procedure permits this choice of reference. They discuss several references, including percentile-based profiles and the point on the decision boundary closest to the declined applicant \cite{nairExplainingAdverseActions2022}, though their illustration uses a favorable profile at the 75th percentile of each predictor. The considerations above therefore justify some ways of filling out the reference in $\Sigma^{\operatorname{Nair}}$. 
 
Turning to the second substantive choice within the comparison distribution, we consider the Shapley kernel $\kappa$. Together with the assessment and correlation aggregation, its weighting of comparisons ranks notices by the mean B-Shap contribution of their factors (Section~\ref{sec:nair-reconstruction}). \citet{nairExplainingAdverseActions2022} motivate B-Shap as a way to allocate the applicant--reference score difference while accounting for interactions among factors. This supplies an explanatory rationale for the ranking: allocating joint effects avoids counting the same effect in full toward several factors' contributions. However, these properties support the ranking only if principal reasons are understood as the factors bearing the largest shares of responsibility for the applicant--reference score difference. Institutional considerations do not require this interpretation. A creditor need not describe how or why a factor adversely affected an applicant (comment 9(b)(2)-3), so the notice identifies reasons without quantifying their contributions \cite{cfpbOfficialInterp10029Notifications}. Likewise, the institutional requirement to disclose the actual principal reasons constrains what the notice may report but does not establish how joint effects should determine disclosure priority \cite{cfpbOfficialInterp10029Notifications}.
 
Normative considerations provide a competing reason. The educational and error-correction purposes of AANs favor drawing attention to consequential deficiencies in the applicant's record. A notice serves these purposes when it directs the applicant to the factors whose correction would most improve this application. Dividing a joint effect can lower a factor's priority for disclosure even when changing that factor alone would remove the entire effect from this application. For instance, suppose a short credit history and high utilization raise the log-odds by $4$ only when they occur together, while delinquency and inquiries raise it by $3$ and $2.5$ independently. B-Shap assigns $2$ to each interacting factor, so a two-reason notice names delinquency and inquiries. Yet correcting either interacting factor alone would lower the score by $4$, more than correcting either named factor. If the utilization entry were misreported, the notice would give the applicant no reason to check it. This counts against the allocation-based ranking insofar as disclosure should help applicants identify consequential changes or errors. Still, this does not establish that individual replacement effects must determine principal reasons. When a penalty applies if either of two factors is adverse, replacing either factor alone leaves the penalty in place, so individual effects credit it to neither factor. The normative consideration therefore identifies a purpose against which the allocation must be justified, without supplying a rule to replace it.
 
The setting, therefore, provides grounds for tying the reference to the standard for approval, particularly when it is close to the decision boundary. For the treatment of interactions, the setting supplies competing considerations: apportioning joint effects supports the Shapley ranking, while directing attention to consequential deficiencies can count against it. Justifying the ranking requires explaining why the former should govern disclosure priority and how the resulting selection serves the latter purpose. The faithfulness functional makes this judgment explicit by locating the comparisons and their treatment through which the notice ranking is determined.

\section{Conclusion}
\label{sec:conclusion}
 
In this work, we have argued that faithfulness names a common, incompletely specified relation between a model and an explanation. Our proposed faithfulness functional makes explicit the choices through which existing evaluations specify that relation. Our reconstructions of a selected family of faithfulness operationalizations show how criteria can differ in what they require of an explanation even within a shared behavioral understanding of faithfulness. The case study of adverse action notices demonstrates how these choices can be examined against the explanation's form, institutional requirements, and purposes.

More broadly, we see this as a further example from the responsible machine learning literature of the problems that arise when measurement modeling is neglected \cite{jacobsMeasurementFairness2021, liuBridgingPredictionIntervention2026, costonValidityPerspectiveEvaluating2023, wallachEvaluatingGenerativeAI2024}. Even a concept like faithfulness, which seems to denote a clear property of an explanation, turns out to be difficult to operationalize in a way that captures the underlying notion. Operationalizations proposed as general measures of faithfulness each commit to a particular reading of what an explanation claims, and the suitability of that commitment for a particular use requires justification. Making it explicit, as the faithfulness functional does, is a precondition for asking whether a given operationalization is valid for the use to which it is put. 

Justifying an operationalization therefore requires attention to considerations that the general requirement of faithfulness does not supply. In the adverse action notice case, these considerations give grounds for some reference choices and expose unresolved questions about the treatment of interactions. The resulting contribution is a way to distinguish supported choices from those requiring further argument. Reporting a faithfulness specification together with its justification makes both available for scrutiny and provides a basis for evaluating whether the resulting judgments are appropriate for the explanation's use.

\section*{Generative AI Usage Statement}
The final version of this text was copy-edited and proofread using AI tools. Generative AI also assisted with and checked proofs. Any remaining errors are ours.

\bibliographystyle{ACM-Reference-Format}
\bibliography{refrences}

\appendix

\section{Rank non-equivalence of example faithfulness operationalizations}
\label{app:non-equivalence}

We consider the three criteria in Section~\ref{sec:operationalizing-faithfulness}. Fix $\mathcal X=\{0,1\}^4$, the explained input $x=(1,1,1,1)$, the baseline $b=(0,0,0,0)$, and $k=2$. Here $x^{\setminus S}$ replaces the features in $S$ by their baseline values, and $x^S$ retains the features in $S$ and replaces the rest. Explanations are attribution vectors $\phi\in\mathbb R^4$, with features ranked by decreasing attribution value; the rationale criteria evaluate $R=S_2(\phi)$. For explanation infidelity, $\mu_I$ is uniform over the coordinate basis vectors $u_1,\ldots,u_4$. Rankings use the relation $\succ_F$ of Definition~\ref{def:rank-non-equivalence}, with attribution vectors as the candidate explanations $e$ and $e'$.

\begin{proposition}
\label{prop:pairwise-non-equivalence}
Explanation infidelity, sufficiency, and comprehensiveness, as defined in Section~\ref{sec:operationalizing-faithfulness}, are pairwise rank-non-equivalent in the sense of Definition~\ref{def:rank-non-equivalence}.
\end{proposition}

\begin{proof}
By Definition~\ref{def:rank-non-equivalence}, it suffices to exhibit, for each pair of criteria, two explanations of the same model output that the criteria rank in opposite orders. The following construction supplies all three reversals using one score $s$, one input $x$, and three attribution vectors.

Let $s$ be given by
{\setlength{\arraycolsep}{4pt}
\[
\begin{array}{c|cccccc|cccc|c}
    z
      & 1100 & 1010 & 1001 & 0110 & 0101 & 0011
      & 0111 & 1011 & 1101 & 1110
      & 1111\\
    \hline
    s(z)
      & 1 & 3 & 5 & 7 & 4 & 6
      & 8 & 17 & 18 & 19
      & 20
\end{array}
\]}
and let $s$ vanish at $0000$ and at the four singleton inputs, whose values never enter the computations below. The vertical rules separate vertices by the number of active features. The map $z\mapsto s(z)$ is coordinatewise monotone, so the reversals do not depend on a nonmonotone score. Sufficiency and comprehensiveness are stated for probability scores. Dividing $s$ and every attribution vector below by $20$ places $s$ in $[0,1]$, so that $s/20$ can be read as the probability a binary classifier assigns to the class it predicts at $x$. This multiplies each criterion by a positive constant, so every ranking below is unchanged.

Reading the last two blocks against $s(x)=20$ gives
\[
    s(x)-s\bigl(x^{\setminus\{j\}}\bigr) = 12,\;3,\;2,\;1 \qquad\text{for }j=1,2,3,4.
\]
Take
\[
    \phi^{\mathrm I}=(12,3,2,1),\quad \phi^{\mathrm S}=(2,4,3,1),\quad \phi^{\mathrm C}=(2,1,4,3),
\]
so that
\[
    S_2(\phi^{\mathrm I})=\{1,2\},\quad S_2(\phi^{\mathrm S})=\{2,3\},\quad S_2(\phi^{\mathrm C})=\{3,4\}.
\]

Perturbing by $u_j$ sets feature $j$ to its baseline value, since $x-u_j=x^{\setminus\{j\}}$. Thus, Equation~\eqref{eq:infidelity} becomes
\[
    \operatorname{INFD}(\phi;s,x,\mu_I) = \frac14\sum_{j=1}^{4} \left( \phi_j- \bigl[s(x)-s(x^{\setminus\{j\}})\bigr] \right)^2.
\]
This is a squared distance, so among the three explanations it is uniquely minimized when $\phi$ matches the four single-feature score drops, namely at $\phi^{\mathrm I}$.

The other two criteria depend on $\phi$ only through $R=S_2(\phi)$, as in Equations~\eqref{eq:sufficiency} and~\eqref{eq:comprehensiveness}. Reading the needed values of $s$ from the table gives
\[
\begin{array}{c|ccc}
      & \operatorname{INFD}(\phi;s,x,\mu_I)
      & \operatorname{SUFF}(S_2(\phi);s,x)
      & \operatorname{COMP}(S_2(\phi);s,x)\\
      & \text{(low)}
      & \text{(low)}
      & \text{(high)}\\
    \hline
    \phi^{\mathrm I} & \mathbf{0} & 19 & 14\\
    \phi^{\mathrm S} & 51/2 & \mathbf{13} & 15\\
    \phi^{\mathrm C} & 28 & 14 & \mathbf{19}
\end{array}
\]
where the parenthetical labels indicate the direction in which the criterion orders explanations as more faithful. The table gives the three strict ranking reversals:
\[
\begin{aligned}
\phi^{\mathrm I}&\succ_{\mathrm{INFD}}\phi^{\mathrm S},
&\qquad \phi^{\mathrm S}&\succ_{\mathrm{SUFF}}\phi^{\mathrm I},\\
\phi^{\mathrm I}&\succ_{\mathrm{INFD}}\phi^{\mathrm C},
&\qquad \phi^{\mathrm C}&\succ_{\mathrm{COMP}}\phi^{\mathrm I},\\
\phi^{\mathrm S}&\succ_{\mathrm{SUFF}}\phi^{\mathrm C},
&\qquad \phi^{\mathrm C}&\succ_{\mathrm{COMP}}\phi^{\mathrm S}.
\end{aligned}
\]
Each row supplies two explanations of the same model output that the corresponding pair of criteria ranks in opposite orders, as required by Definition~\ref{def:rank-non-equivalence}. Hence the three criteria are pairwise rank-non-equivalent.
\end{proof}

\section{Translations into faithfulness specifications}
\label{app:translation-long-notes}

This appendix verifies the eight reconstructions summarized in Table~\ref{tab:unified-faithfulness}. For each operationalization, we first clarify why it falls within the sense of faithfulness considered in this paper. Some source papers explicitly present their criteria as measures of faithfulness or fidelity; others use different terminology but evaluate whether an explanation accurately represents the model behavior it purports to describe. We then give the specification $\Sigma=\langle\nu_{x,e},\ell_{x,e},\mathcal A\rangle$ and show that it recovers the source criterion. As in Section~\ref{sec:uff}, metric superscripts identify a specification and its three components; the comparison law, assessment, and aggregation defined in each subsection together specify the corresponding $\Sigma$. Conditions (i)--(iv) refer to Definition~\ref{def:admissible}.

We use the notation of Sections~\ref{sec:operationalizing-faithfulness}--\ref{sec:uff}: $f$ is the prediction model, $s$ is the scalar score being assessed, and $e$ is an attribution vector $\phi$ or a selected rationale $R$. As in Section~\ref{sec:operationalizing-faithfulness}, for a classifier $s(z)=f_c(z)$, where $c=\arg\max_j f_j(x)$ is the class predicted for the explained input and is held fixed throughout each evaluation. The set $S_k(\phi)$ contains the $k$ highest-attributed features, with $S_0(\phi)=\varnothing$ and ties resolved by a fixed rule. A rationale can be supplied directly or selected as $R=S_k(\phi)$. The operations $x^{\setminus S}$ and $x^S$ follow each source's procedure: faithfulness correlation replaces selected coordinates with baseline values, comprehensiveness and sufficiency delete or retain selected tokens, and insertion restores pixels onto a blurred copy of $x$. Random perturbations are specified by $q(\cdot\mid x,S)$, the distribution over inputs obtained by perturbing the features in $S$ while holding the rest fixed.

For curve-based criteria, $S\sim\pi_\phi$ is the draw in Table~\ref{tab:unified-faithfulness}: $\pi_\phi$ assigns each evaluated set $S_k(\phi)$ its normalized averaging or integration weight. The evaluated feature counts and weights are fixed across explanations of the same input; the selected sets are determined by $\phi$. Each subsection defines $\nu_{x,e}$ as the law of its comparison $\omega$. As in Section~\ref{sec:uff}, $x'$ abbreviates the comparison point $x'_\omega$ determined by $\omega$.

We retain the source-score directions of Section~\ref{sec:operationalizing-faithfulness} and orient each reconstruction so that larger values indicate greater faithfulness. Thus, the $F(e;x,s)$ reconstructed in Section~\ref{sec:uff} is the source criterion after reversing its sign where lower values are preferred. Unless stated otherwise, $a_{x,s}=1$ and $b_{x,s}=0$. If a source curve score has averaging or integration weights summing to $W>0$, $\pi_\phi$ divides those weights by $W$, giving $a_{x,s}=1/W$ and $b_{x,s}=0$.

\subsection{Explanation infidelity}
\label{app:translation-infidelity}

\citet{yehInfidelitySensitivityExplanations2019} explicitly introduce explanation infidelity as an objective measure of the fidelity of an explanation to a predictor. The attribution vector is interpreted as predicting the model-score change associated with a meaningful input perturbation, and infidelity measures the discrepancy between the predicted and observed changes. The source measure is
\begin{equation}
\label{eq:app-original-infidelity}
\operatorname{INFD}(\phi;s,x,\mu_I) = \mathbb E_{I\sim\mu_I} \left[ \left( \phi^\top I-\left[s(x)-s(x-I)\right] \right)^2 \right].
\end{equation}
Let $\mu_I$ be the specified distribution of input perturbations, with $x-I\in\mathcal X$. For $e=\phi$, let $\nu^{\operatorname{INFD}}_{x,\phi}$ be the law of $\omega=I\sim\mu_I$, with $x'=x-I$, and define
\[
\ell^{\operatorname{INFD}}_{x,\phi}(\omega;s) = -\Bigl(s(x)-s(x')-\phi^\top(x-x')\Bigr)^2,
\qquad
\mathcal A^{\operatorname{INFD}}=\mathbb E .
\]
The explanation enters the assessment rule, supplying the predicted change. Because $x-x'=I$,
\[
\operatorname{Faith}_{\Sigma^{\operatorname{INFD}}}(\phi;x,s) = -\mathbb E_{I\sim\mu_I} \left[ \left(\phi^\top I-\left[s(x)-s(x-I)\right]\right)^2 \right] = -\operatorname{INFD}(\phi;s,x,\mu_I),
\]
reversing the source orientation, under which lower infidelity indicates greater faithfulness. The specified perturbation law generates comparisons without additional queries to $s$, satisfying (i); the assessment consults $s$ at $x$ and $x'$ alone and varies with both outputs, satisfying (ii); expectation depends only on the distribution of scalar assessments and is monotone, satisfying (iii); and the explanation enters through $\ell^{\operatorname{INFD}}_{x,\phi}$, which supplies the predicted change, satisfying (iv).

\subsection{Faithfulness correlation}
\label{app:translation-faithfulness-correlation}

\citet{3491440.3491857} explicitly define this criterion as the faithfulness of a feature-based explanation to a predictor at a particular input. Their motivating claim is that the attribution assigned to a feature subset should vary proportionally with the change in the model output when that subset is replaced by baseline values.

Fix a subset size $n$ and a baseline $b$. For $e=\phi$, let $\nu^{\operatorname{FC}}_{x,\phi}$ be the law of $\omega=S\sim\operatorname{Unif}\{T\subseteq[d]:|T|=n\}$, with $x'=x^{\setminus S}$ replacing coordinates in $S$ by their values in $b$, and define
\[
\ell^{\operatorname{FC}}_{x,\phi}(\omega;s)
=
\Bigl(s(x)-s(x'),\;\textstyle\sum_{j\in S}\phi_j\Bigr),
\qquad
\mathcal A^{\operatorname{FC}}=\operatorname{Corr}.
\]
Then
\[
\operatorname{Faith}_{\Sigma^{\operatorname{FC}}}(\phi;x,s)
=
\operatorname{Corr}_{S}
\Bigl(
s(x)-s\bigl(x^{\setminus S}\bigr),\;\textstyle\sum_{j\in S}\phi_j
\Bigr),
\]
which is the source's Definition~3, with Pearson correlation defined when both coordinates have positive variance. Subset selection uses the declared baseline without querying $s$, satisfying (i). The assessment pairs the observed score drop with the attribution sum, satisfying (ii) and (iv); correlation depends only on the distribution of these pairs, satisfying (iii).

\subsection{Comprehensiveness}
\label{app:translation-comprehensiveness}

\citet{deyoungERASERBenchmarkEvaluate2020} explicitly introduce comprehensiveness as a faithfulness scoring metric for rationales. A rationale is interpreted as containing information that meaningfully influenced the model's prediction, and the criterion tests this by deleting the rationale tokens and measuring the resulting reduction in the original predicted-class score. The class is held fixed across the comparison. The source criterion is
\begin{equation}
\label{eq:app-original-comprehensiveness}
\operatorname{COMP}(R;s,x)=s(x)-s\bigl(x^{\setminus R}\bigr).
\end{equation}
For $e=R$, let $\nu^{\operatorname{COMP}}_{x,R}$ place all mass on $\omega=R$, with $x'=x^{\setminus R}$, and define
\[
\ell^{\operatorname{COMP}}_{x,R}(\omega;s)=s(x)-s(x'),
\qquad
\mathcal A^{\operatorname{COMP}}=\mathbb E .
\]
Thus,
\[
\operatorname{Faith}_{\Sigma^{\operatorname{COMP}}}(R;x,s)
=
s(x)-s\bigl(x^{\setminus R}\bigr)
=
\operatorname{COMP}(R;s,x).
\]
The comparison point is constructed from the rationale without consulting $s$, satisfying (i); the assessment consults $s$ at $x$ and $x'$ alone and varies with both outputs, satisfying (ii); expectation is monotone and depends only on the assessment distribution, satisfying (iii); and here the explanation enters through $\nu^{\operatorname{COMP}}_{x,R}$ rather than $\ell^{\operatorname{COMP}}_{x,R}$, since the assessment is the same function of the two scores for every explanation of the same output, satisfying (iv).

\subsection{Area over the perturbation curve}
\label{app:translation-aopc}

\citet{samekEvaluatingVisualizationWhat2017} present AOPC as an objective quality measure for heatmaps intended to explain what made a neural network reach a particular classification decision. The heatmap is interpreted as ranking input regions by their relevance to that decision, and the ranking is evaluated by measuring how quickly the model score declines when regions are perturbed in that order. This falls within our account of faithfulness because it tests whether the ordering represented by the explanation agrees with the model's behavior under the corresponding perturbations. In the source's principal experiments, the ranked units are nonoverlapping $9\times9$ image regions; here $[d]$ indexes those regions and $S_k(\phi)$ selects the $k$ highest-ranked regions. Fix the number of perturbation steps $L\in\{1,\ldots,d\}$. Let $q(\cdot\mid x,S)$ replace the pixels in the selected regions with uniformly sampled values. Averaging over these replacements gives the per-input score
\begin{equation}
\label{eq:app-original-aopc}
\operatorname{AOPC}_L(\phi;s,x)
=
\frac{1}{L+1}\sum_{k=0}^{L}
\mathbb E_{x'\sim q(\cdot\mid x,S_k(\phi))}
\left[s(x)-s(x')\right].
\end{equation}
Let $\pi_\phi$ assign probability $1/(L+1)$ to each prefix $S_k(\phi)$ for $k=0,\ldots,L$. For $e=\phi$, let $\nu^{\operatorname{AOPC}}_{x,\phi}$ be the joint law of $\omega=(S,x')$ obtained by drawing $S\sim\pi_\phi$ and $x'\sim q(\cdot\mid x,S)$, and define
\[
\ell^{\operatorname{AOPC}}_{x,\phi}(\omega;s)=s(x)-s(x'),
\qquad
\mathcal A^{\operatorname{AOPC}}=\mathbb E .
\]
Then
\[
\operatorname{Faith}_{\Sigma^{\operatorname{AOPC}}}(\phi;x,s)
=
\frac{1}{L+1}\sum_{k=0}^{L}
\mathbb E_{x'\sim q(\cdot\mid x,S_k(\phi))}
\left[s(x)-s(x')\right]
=
\operatorname{AOPC}_L(\phi;s,x).
\]
This specification shares its assessment rule with comprehensiveness and differs only in which comparisons are made. Averaging these per-input scores over the evaluation images gives the source's dataset-level score. The original ranking determines the prefixes, and the replacement draws are independent of further model responses, satisfying (i); the assessment and pooling are that of comprehensiveness and satisfy (ii) and (iii) for the same reasons; and the explanation again enters through $\nu^{\operatorname{AOPC}}_{x,\phi}$, satisfying (iv).

\citet{deyoungERASERBenchmarkEvaluate2020} also average comprehensiveness and sufficiency across rationale lengths based on the top $1\%,5\%,10\%,20\%$, and $50\%$ of tokens. These variants use token deletion or retention with the corresponding assessment rule, and their evaluated threshold positions and averaging weights determine $\pi_\phi$.

\subsection{Prediction gap on important features}
\label{app:translation-pgi}

\citet{agarwalOpenXAITransparentEvaluation2022} explicitly classify PGI as a predictive-faithfulness metric. It evaluates whether the features identified as important by an attribution explanation are consequential to the underlying model by perturbing those features and testing whether the prediction probability changes substantially. For continuous features, $q(\cdot\mid x,S)$ is the distribution of $x+\mathbf 1_S\odot\varepsilon$, where $\varepsilon$ follows the specified Gaussian noise law and $\mathbf 1_S$ is the coordinate mask. For fixed $k$ the source criterion is
\begin{equation}
\label{eq:app-original-pgi}
\operatorname{PGI}_k(\phi;s,x)
=
\mathbb E_{x'\sim q(\cdot\mid x,S_k(\phi))}
\bigl[\left|s(x)-s(x')\right|\bigr],
\end{equation}
and the reported metric is the area under this quantity as $k$ varies. For $e=\phi$, let $\nu^{\operatorname{PGI}}_{x,\phi}$ be the joint law of $\omega=(S,x')$ obtained by drawing $S\sim\pi_\phi$ and $x'\sim q(\cdot\mid x,S)$, and define
\[
\ell^{\operatorname{PGI}}_{x,\phi}(\omega;s)=\bigl|s(x)-s(x')\bigr|,
\qquad
\mathcal A^{\operatorname{PGI}}=\mathbb E .
\]
Then
\[
\operatorname{Faith}_{\Sigma^{\operatorname{PGI}}}(\phi;x,s)
=
\mathbb E_{S\sim\pi_\phi}
\left[
\mathbb E_{x'\sim q(\cdot\mid x,S)}
\bigl[\left|s(x)-s(x')\right|\bigr]
\right],
\]
the normalized PGI area under the curve; an implementation using an unnormalized area differs by a fixed positive factor. The joint comparison law is specified by the attribution ranking and the declared noise distribution without additional queries to $s$, satisfying (i); the assessment consults $s$ at $x$ and $x'$ alone and varies with both outputs, satisfying (ii); the pooling is expectation again, satisfying (iii); and the explanation enters through $\nu^{\operatorname{PGI}}_{x,\phi}$, which perturbs the features it ranks highest, satisfying (iv).

\subsection{Sufficiency}
\label{app:translation-sufficiency}

\citet{deyoungERASERBenchmarkEvaluate2020} explicitly introduce sufficiency as a faithfulness scoring metric for rationales. A selected rationale is treated as claiming to contain enough of the information used by the model to reach its original disposition, and the criterion tests that claim by retaining only the rationale tokens and measuring the reduction in the original predicted-class score. The source criterion is
\begin{equation}
\label{eq:app-original-sufficiency}
\operatorname{SUFF}(R;s,x)=s(x)-s\bigl(x^{R}\bigr),
\end{equation}
for which lower values are preferred. For $e=R$, let $\nu^{\operatorname{SUFF}}_{x,R}$ place all mass on $\omega=R$, with $x'=x^{R}$, and define
\[
\ell^{\operatorname{SUFF}}_{x,R}(\omega;s)=s(x')-s(x),
\qquad
\mathcal A^{\operatorname{SUFF}}=\mathbb E ,
\]
so that
\[
\operatorname{Faith}_{\Sigma^{\operatorname{SUFF}}}(R;x,s)
=
s\bigl(x^{R}\bigr)-s(x)
=
-\operatorname{SUFF}(R;s,x).
\]
The specification recovers the source criterion with its orientation reversed, and rewards a high original-class score on the retained input. The retained input is constructed from the rationale without consulting $s$, satisfying (i); the assessment consults $s$ at $x$ and $x'$ alone and varies with both outputs, satisfying (ii); expectation satisfies (iii); and the explanation enters through $\nu^{\operatorname{SUFF}}_{x,R}$, satisfying (iv).

\subsection{Insertion AUC}
\label{app:translation-insertion}

\citet{petsiukRISERandomizedInput2018} introduce insertion and deletion as automatic causal metrics for saliency explanations. Insertion interprets the saliency ordering as identifying the features whose presence supports the model output: starting from the blurred image $b=\operatorname{Blur}(x)$ and restoring batches of pixels in that order, earlier prefixes should rapidly recover the original class score. This falls within our account because it tests whether restoring the features the explanation identifies reproduces the model's original response. Here $x^S$ takes its pixels in $S$ from $x$ and the remaining pixels from $b$. Let $\pi_\phi$ assign the normalized trapezoidal integration weights to the sets restored at the evaluated steps, including $S_0(\phi)=\varnothing$ with score $s(b)$. The insertion score is
\begin{equation}
\label{eq:app-original-insertion}
\operatorname{InsAUC}(\phi;s,x)=\mathbb E_{S\sim\pi_\phi}\bigl[s(x^S)\bigr].
\end{equation}
For $e=\phi$, let $\nu^{\operatorname{InsAUC}}_{x,\phi}$ be the law of $\omega=S\sim\pi_\phi$, with $x'=x^S$, and define
\[
\ell^{\operatorname{InsAUC}}_{x,\phi}(\omega;s)=s(x'),
\qquad
\mathcal A^{\operatorname{InsAUC}}=\mathbb E .
\]
Then
\[
\operatorname{Faith}_{\Sigma^{\operatorname{InsAUC}}}(\phi;x,s)
=
\mathbb E_{S\sim\pi_\phi}\bigl[s(x^S)\bigr]
=
\operatorname{InsAUC}(\phi;s,x),
\]
recovering the insertion AUC. The blurred baseline and evaluated pixel counts are fixed independently of the explanation. The restored prefixes are determined by the ranking without consulting $s$, satisfying (i); the assessment depends on $s(x')$, satisfying (ii), and expectation satisfies (iii); and the explanation enters through $\nu^{\operatorname{InsAUC}}_{x,\phi}$, satisfying (iv).

The source's deletion AUC, $\operatorname{DelAUC}(\phi;s,x)$, sets ranked pixels to zero and integrates the remaining original-class score, with smaller areas preferred. The same construction with $S\sim\pi_\phi$, $x'=x^{\setminus S}$, and assessment $-s(x')$ recovers its negative; the score-drop assessment gives the equivalent value $s(x)-\operatorname{DelAUC}(\phi;s,x)$ for normalized weights.

\subsection{Prediction gap on unimportant features}
\label{app:translation-pgu}

\citet{agarwalOpenXAITransparentEvaluation2022} explicitly classify PGU as the complementary predictive-faithfulness metric to PGI. It tests whether features treated as unimportant are in fact inconsequential to the model by perturbing them and testing whether the prediction probability is left approximately unchanged. Using the same perturbation law $q$ as PGI, for fixed $k$ the source measure is
\begin{equation}
\label{eq:app-original-pgu}
\operatorname{PGU}_k(\phi;s,x)
=
\mathbb E_{x'\sim q(\cdot\mid x,[d]\setminus S_k(\phi))}
\bigl[\left|s(x)-s(x')\right|\bigr],
\end{equation}
for which lower values are preferred. For $e=\phi$, let $\nu^{\operatorname{PGU}}_{x,\phi}$ be the joint law of $\omega=(S,x')$ obtained by drawing $S\sim\pi_\phi$ and $x'\sim q(\cdot\mid x,[d]\setminus S)$, and define
\[
\ell^{\operatorname{PGU}}_{x,\phi}(\omega;s)=-\bigl|s(x)-s(x')\bigr|,
\qquad
\mathcal A^{\operatorname{PGU}}=\mathbb E ,
\]
so that
\[
\operatorname{Faith}_{\Sigma^{\operatorname{PGU}}}(\phi;x,s)
=
-\mathbb E_{S\sim\pi_\phi}
\left[
\mathbb E_{x'\sim q(\cdot\mid x,[d]\setminus S)}
\bigl[\left|s(x)-s(x')\right|\bigr]
\right],
\]
the negative normalized PGU area under the curve, reversing the source orientation. This specification shares a comparison-generating scheme with PGI, differing in which features are perturbed and in the sign of the assessment. Admissibility holds as for PGI, with the explanation entering $\nu^{\operatorname{PGU}}_{x,\phi}$ through the complement of its highest-ranked set.

These reconstructions differ in which features they select, how they alter those features, what model response they reward, and how they combine comparisons. Comprehensiveness and AOPC reward signed score drops, PGI rewards absolute changes, and PGU penalizes those changes on the complementary features; sufficiency and insertion reward high scores on retained or restored inputs. Comprehensiveness, AOPC, sufficiency, and insertion therefore rely on the orientation of $s$ fixed in Section~\ref{sec:operationalizing-faithfulness}, since they reward score changes or levels in one direction. PGI and PGU, which use absolute changes, do not.

\section{Reconstruction of B-Shap reason selection}
\label{app:aan-correlation}

Use the setting and notation of Definition~\ref{def:nair}, with all
comparison inputs in the model's domain. Write $\phi_g$ for factor
$g$'s B-Shap contribution and define
\[
F(R_N;x,s)=\frac{1}{k}\sum_{g\in N}\phi_g.
\]
We show that the correlation score in
Equation~\eqref{eq:aan-correlation} is a positive affine
transformation of $F$, with constants independent of the notice.
Its maximizing notices therefore name the $k$ largest B-Shap
contributions, allowing for ties.

\paragraph{Rewriting the contributions.}
Let $\Delta(D)=s(x)-s(x^{\setminus D})$. Reversing the order in
which factors switch from reference to applicant values expresses
B-Shap in terms of these score drops:
\[
\phi_g
=
\sum_{D\subseteq\mathcal G\setminus\{g\}}
w_{|D|}
\bigl[\Delta(D\cup\{g\})-\Delta(D)\bigr],
\qquad
w_t=\frac{t!(m-t-1)!}{m!}.
\]
For a nonempty proper subset $D$ of size $t$, the coefficient of
$\Delta(D)$ is $w_{t-1}$ when $g\in D$, and $-w_t$ otherwise.
Equivalently, it is
\[
\mathbf 1_{\{g\in D\}}(w_{t-1}+w_t)-w_t,
\qquad
w_{t-1}+w_t
=
\frac{m}{\binom mt\,t(m-t)}.
\]
The last expression is proportional to $\kappa(D)$. The remaining
coefficient $-w_t$ does not depend on $g$, nor does the contribution
from the full set; the empty set contributes zero. Consequently,
\[
\phi_g
=
c\,\mathbb E_{D\sim\kappa}
\bigl[\Delta(D)\mathbf 1_{\{g\in D\}}\bigr]+C,
\]
for constants $c>0$ and $C$ that are the same for every factor.
Their exact values are not needed.

\paragraph{Recovering the notice score.}
Averaging this identity over the $k$ named factors gives
\[
F(R_N;x,s)
=
c\,\mathbb E_{D\sim\kappa}
\left[\Delta(D)\frac{|N\cap D|}{k}\right]+C.
\]
Because $\kappa$ is unchanged by relabeling factors, the fraction
$|N\cap D|/k$ has the same distribution for every notice of length
$k$. Its variance is positive: singleton subsets can contain either
a named or an unnamed factor, since $1\le k<m$. The distribution
of $\Delta(D)$ is also independent of the notice.

Provided $\Delta(D)$ has positive variance, converting the expected
product above into a covariance subtracts the same quantity for
every notice, and converting that covariance into a correlation
divides by the same positive quantity. Thus,
\[
\operatorname{Faith}_{\Sigma^{\operatorname{Nair}}}(R_N;x,s)
=
\operatorname{Corr}_{D\sim\kappa}
\left(\Delta(D),\frac{|N\cap D|}{k}\right)
\]
is a positive affine transformation of $F$, preserving all rankings
and ties as required by Section~\ref{sec:uff}.

If $\Delta(D)$ has zero variance, it is constant on the comparison
subsets. Symmetry then makes the rewritten contribution identical
for every factor, so all notices tie. Assigning every notice a score
of zero extends the otherwise undefined correlation while preserving
this tie.

Finally, the specification satisfies
Definition~\ref{def:admissible}: comparisons use the fixed reference
and partition without further model queries (i); the assessment
depends on the model only through the compared scores (ii);
correlation depends only on the distribution of assessment pairs
(iii); and the second coordinate depends on the notice (iv).

\end{document}